\documentclass[acmsmall]{acmart}
\usepackage{amsmath}
\usepackage{bm}
\usepackage{algorithm}
\usepackage{algpseudocode}
\usepackage{amsmath}
\usepackage{multirow}
\usepackage{indentfirst}
\usepackage{subfigure}
\usepackage{epsfig}
\usepackage{epstopdf}
\usepackage{graphicx}
\usepackage{url}
\usepackage{xspace}
\usepackage{booktabs}
\usepackage{subeqnarray}
\usepackage{subcaption}
\usepackage{color}

\newcommand{\ignore}[1]{}

\usepackage{makecell}
\AtBeginDocument{%
  }

\acmJournal{TOIS}

\begin{document}

\title{ArbGraph: Conflict-Aware Evidence Arbitration for Reliable Long-Form Retrieval-Augmented Generation}


\author{Qingying Niu}
\authornote{Equal Contribution.}
\email{qingyingniu36@gmail.com}
\affiliation{%
  \department[0]{Gaoling School of Artificial Intelligence}
  \institution{Renmin University of China}
  \city{Beijing}
  \country{China}
  \postcode{100872}
}

\author{Yuhao Wang}
\authornotemark[1]
\email{yh.wang@ruc.edu.cn}
\affiliation{%
  \department[0]{Gaoling School of Artificial Intelligence}
  \institution{Renmin University of China}
  \city{Beijing}
  \country{China}
  \postcode{100872}
}

\author{Ruiyang Ren}
\authornote{Corresponding Authors.}
\email{reyon_ren@outlook.com}
\affiliation{%
  \department[0]{Gaoling School of Artificial Intelligence}
  \institution{Renmin University of China}
  \city{Beijing}
  \country{China}
  \postcode{100872}
}

\author{Bohui Fang}
\email{bohuif@alumni.cmu.edu}
\affiliation{%
  \institution{Carnegie Mellon University}
  \city{Pittsburgh}
  \state{Pennsylvania}
  \country{USA}
  \postcode{15213}
}


\author{Wayne Xin Zhao}
\authornotemark[2]
\email{batmanfly@gmail.com}
\affiliation{%
  \department[0]{Gaoling School of Artificial Intelligence}
  \institution{Renmin University of China}
  \city{Beijing}
  \country{China}
  \postcode{100872}
}






\renewcommand{\shortauthors}{Niu et al.}

\begin{abstract}
Retrieval-augmented generation (RAG) remains unreliable in long-form settings, where retrieved evidence is noisy or contradictory, making it difficult for RAG pipelines to maintain factual consistency. Existing approaches focus on retrieval expansion or verification during generation, leaving conflict resolution entangled with generation.
To address this limitation, we propose ArbGraph, a framework for pre-generation evidence arbitration in long-form RAG that explicitly resolves factual conflicts. ArbGraph decomposes retrieved documents into atomic claims and organizes them into a conflict-aware evidence graph with explicit support and contradiction relations. On top of this graph, we introduce an intensity-driven iterative arbitration mechanism that propagates credibility signals through evidence interactions, enabling the system to suppress unreliable and inconsistent claims before final generation. In this way, ArbGraph separates evidence validation from text generation and provides a coherent evidence foundation for downstream long-form generation.
We evaluate ArbGraph on two widely used long-form RAG benchmarks, LongFact and RAGChecker, using multiple large language model backbones. Experimental results show that ArbGraph consistently improves factual recall and information density while reducing hallucinations and sensitivity to retrieval noise. Additional analyses show that these gains are evident under conflicting or ambiguous evidence, highlighting the effectiveness of evidence-level conflict resolution for improving the reliability of long-form RAG. The implementation is publicly available at https://github.com/1212Judy/ArbGraph.
\end{abstract}



\maketitle

\section{Introduction}

Retrieval-Augmented Generation (RAG) has emerged as a widely used paradigm for grounding large language models in external knowledge~\cite{lewis2020rag, gao2024rag_survey, rocketqa, ren2025investigating}. However, its reliability depends not only on whether relevant evidence can be retrieved, but also on whether that evidence can be consolidated into a coherent factual basis for generation~\cite{yoran2023robust}. These challenges are especially pronounced in long-form settings, where models must synthesize multiple interdependent facts into extended responses rather than produce isolated short answers~\cite{liu2023lost_middle}. In such contexts, factual errors rarely remain local. Instead, noisy, redundant or mutually inconsistent evidence can distort the evolving discourse structure, allowing early mistakes to propagate across subsequent claims and ultimately undermine global factual coherence~\cite{wei2024longform,huang2024calibrating}. Simply increasing retrieval depth does not necessarily improve reliability, since additional retrieved evidence often introduces further inconsistencies rather than greater certainty that complicate downstream reasoning.


Typically, attempts to improve long-form factuality have largely proceeded along two lines. One seeks to make generation itself more reliable, through self-evaluation, iterative refinement or agentic reasoning that attempts to detect and correct errors as the answer unfolds~\cite{asai2023self, yan2024crag, shinn2023reflexion, xie2025fire, hu2025mctsr}. The other makes evidence more structured, organizing retrieved content into semantic or entity-level representations that improve retrieval coverage and reasoning connectivity~\cite{besta2024got, edge2024graphrag, jimenezgutierrez2024hipporag, zhao2024tapera}. Both directions have yielded important gains, yet both stop short of resolving the same core problem: they do not explicitly determine the credibility of competing evidence before synthesis. Consequently, factual conflict is handled either implicitly during decoding or only indirectly through structural organization, rather than through a direct decision over which claims should be trusted. Under noisy or contradictory retrieval, this limitation becomes critical. Generation-time correction acts only after potentially unreliable evidence has already shaped the response, whereas structural connectivity alone cannot determine which of several plausible but incompatible claims is correct. As a result, although both strategies improve robustness, neither fully addresses the challenge of maintaining factual consistency in long-form generation when retrieved evidence is noisy or contradictory.


We argue that this limitation reflects a deeper mismatch in how long-form RAG currently handles evidence conflict. During generation, decoding is driven primarily by token-level continuation, whereas resolving factual conflict requires explicit comparison among competing pieces of evidence under a broader view of their consistency and support. These two processes are not naturally aligned. When retrieved evidence is noisy or contradictory, claims that are linguistically salient, locally plausible or prominently positioned in context may influence generation even if they are only weakly supported. Moreover, once such claims are incorporated early, subsequent decoding can become conditioned on them, allowing local errors to propagate across the emerging discourse. In long-form settings, where multiple interdependent claims must be integrated into a coherent response, this mismatch becomes particularly consequential. These observations motivate a different strategy: rather than relying primarily on implicit conflict handling during decoding, evidence credibility should be assessed and competing claims should be resolved before generation begins.

To address these challenges, we propose \textbf{ArbGraph}, a framework for pre-generation evidence arbitration in long-form RAG. Rather than leaving conflict handling to the generation process, ArbGraph introduces an explicit arbitration stage between retrieval and synthesis, where competing pieces of evidence are evaluated before they shape the final response. Specifically, ArbGraph decomposes retrieved content into fine-grained \textit{atomic claims}, so that reasoning can be performed over explicit factual units rather than coarse passages.  It then organizes these claims into a conflict-aware evidence graph with explicit \emph{support} and \emph{contradiction} relations. On top of this structure, ArbGraph performs iterative arbitration to reinforce mutually supported claims while suppressing claims that are weakly grounded or mutually inconsistent, yielding a more reliable evidence basis for downstream generation. This process can be viewed as a form of structured evidence filtering, where the credibility of each claim is not determined in isolation but emerges from its interactions with other claims in the graph. The graph serves as an intermediate representation to encode relationships, propagate constraints, and support conflict-aware reasoning.
By shifting conflict handling from the generation stage to the evidence level, ArbGraph transforms evidence selection from a local, token-driven process into a global, structure-aware decision problem. This allows conflict resolution to operate over a complete and consistent view of evidence, reducing sensitivity to prompt structure, attention dynamics, and early decoding biases. As a result, ArbGraph provides a more interpretable and stable foundation for long-form generation under noisy and contradictory retrieval conditions.

Experimental results demonstrate that ArbGraph consistently improves factual recall, faithfulness, and information density, while reducing hallucination and sensitivity to retrieval noise. Further analysis shows that the proposed arbitration mechanism is particularly effective in scenarios involving semantically similar but conflicting evidence or entity ambiguity, where local similarity signals are often insufficient to disambiguate correctness. Overall, these findings indicate that explicit evidence-level arbitration is a promising direction for improving the reliability of long-form RAG under conflicting retrieval conditions.

Our main contributions are summarized as follows:
\begin{itemize}
    \item \textbf{Evidence-level arbitration paradigm:} 
    We formulate evidence arbitration before generation as a distinct reliability mechanism for long-form RAG that moves conflict handling from implicit decoding-time integration to explicit evidence-level decision-making, providing a more stable foundation for downstream generation.

    \item \textbf{Conflict-aware graph modeling:} 
    We construct a structured evidence graph over atomic claims with explicit support and contradiction relations, enabling conflict-aware reasoning under structural constraints. The graph serves as an intermediate representation to support arbitration.

    \item \textbf{Improved robustness in long-form generation:} 
    We conduct extensive experiments on LongFact and RAGChecker across multiple backbones, showing that ArbGraph improves factual robustness and grounding quality, particularly under noisy and conflicting retrieval settings.
    
\end{itemize}
\section{Related Work}
\label{sec:related}

In this section, we review related work from three perspectives: long-form question answering (LFQA), trustworthy RAG under conflicting evidence, and graph-based evidence modeling and reasoning.

\subsection{Long-Form Question Answering (LFQA)}
LFQA research has evolved from early abstractive methods~\cite{fan2019eli5} to retrieval-augmented frameworks that simulate human information seeking and integration, such as WebGPT~\cite{nakano2021webgpt} and WebCPM~\cite{qin2023webcpm}.
To improve verifiability, recent approaches have introduced citation-based supervision~\cite{menick2022teaching}, post-hoc attribution mechanisms~\cite{gao2023rarr}, and intent-aware attribution strategies for long-form answering~\cite{zhao2026improving}.
However, recent benchmarks show that LFQA systems remain vulnerable to noisy retrieval~\cite{pradeep2025nugget, han2024ragqa}, often producing globally inconsistent narratives even when local citations are correct.
These findings suggest that relying solely on source attribution is insufficient for ensuring consistency in long-form generation, as attribution signals alone do not capture interactions or conflicts among multiple pieces of evidence~\cite{pradeep2025nugget, han2024ragqa, zhao2026improving}.
Consequently, improving long-form factuality requires explicitly modeling relationships among evidence and resolving inconsistencies across them.
Our work addresses this limitation by performing explicit evidence-level reasoning and resolving factual conflicts \textit{prior to} generation, rather than relying solely on attribution signals during or after generation.
\subsection{Trustworthy RAG with Conflicting Evidence}
A major challenge for trustworthy RAG lies in handling knowledge conflicts, where retrieved documents may contradict each other or the model’s internal knowledge. This issue is particularly harmful in long-form generation, as errors in key claims can propagate and disrupt the entire output~\cite{wei2024longform}.
Furthermore, irrelevant yet plausible context may distract the model, leading to hallucinations or misinterpretation of evidence~\cite{yoran2023robust,ge2025resolving}.

Existing approaches attempt to mitigate these issues through generation-time or post-hoc mechanisms, including evidence-aware generation, source-level reliability estimation~\cite{hwang2025sourcereliability, ren2025llm, wang2026bee}, and evidence-driven methods~\cite{yue2024evidence}, as well as verification-based methods such as Self-RAG~\cite{asai2023self}, Chain-of-Verification~\cite{dhuliawala2024cove}, Chain-of-Note~\cite{yu-etal-2024-chain}, and TAPERA~\cite{zhao2024tapera}.
In addition, recent works explore claim-level verification or falsification strategies, such as RAC~\cite{li2024rac} and FVA-RAG~\cite{ravishankara2025fva}, as well as fine-tuning approaches for improving reliability~\cite{lee2025finetunerag, wang2025reinforced}.
While these methods improve robustness in certain scenarios, they share a common limitation: conflict handling is primarily performed during or after generation, often relying on heuristic strategies or the model’s implicit reasoning ability.

As a result, conflict resolution remains entangled with the generation process and is sensitive to prompt structure, attention allocation, and early decoding decisions, leading to instability in long-horizon reasoning scenarios~\cite{valmeekam2023planbench}.
Moreover, these approaches typically lack an explicit representation of evidence credibility and do not perform structured evaluation over competing evidence.
In contrast, our approach addresses this limitation by explicitly modeling and resolving conflicts at the evidence level prior to generation, thereby decoupling conflict handling from text synthesis and reducing reliance on implicit and often unstable generation-time reasoning.

\subsection{Graph-based Evidence Modeling and Arbitration}
Graph structures have been widely adopted to improve information organization and reasoning in RAG systems, primarily for enhancing retrieval coverage and connectivity.
Representative methods such as GraphRAG~\cite{edge2024graphrag} and HippoRAG~\cite{jimenezgutierrez2024hipporag} construct entity or semantic graphs to improve information propagation and multi-hop retrieval.
Meanwhile, reasoning-oriented frameworks such as Graph of Thoughts (GoT)~\cite{besta2024got} and Think-on-Graph~\cite{sun2023think} leverage graph topologies to structure inference, improving interpretability and faithfulness~\cite{luo2023reasoning}.

Recent work has also explored resolving factual conflicts via structured representations prior to generation, including argumentation-based approaches~\cite{zhu2025argrag} and knowledge-graph-based conflict resolution methods~\cite{liu2025truthfulrag}.
These studies highlight the potential of graph-structured representations for modeling complex evidence interactions and supporting more structured reasoning.

However, most existing methods use graphs primarily for retrieval expansion or structural organization, rather than as a mechanism for explicit conflict modeling and resolution.
In particular, they often lack explicit representations of contradiction edges and do not perform systematic evaluation or arbitration among conflicting evidence nodes~\cite{edge2024graphrag, jimenezgutierrez2024hipporag, zhu2025argrag, liu2025truthfulrag}.
As a result, these approaches do not explicitly decide between competing claims and are limited in their ability to directly resolve inconsistencies in retrieved evidence.

In contrast, our work treats the graph as a computational framework for conflict-aware arbitration, explicitly modeling support and contradiction relations and enabling structured decision-making over competing evidence prior to generation.
Rather than using graphs solely for connectivity or propagation, ArbGraph leverages them to represent and resolve conflicts, producing a globally consistent and reliable evidence foundation for long-form generation.
\section{Methodology}
\label{sec:method}

In this section, we present the overall methodology of ArbGraph. We first introduce the long-form RAG setting and provide an overview of the proposed framework. We then describe its three core components: atomic claim extraction and semantic alignment, evidence graph construction, and iterative credibility arbitration.

\subsection{Framework Overview}

We consider the long-form retrieval-augmented generation (RAG) setting, where a retriever selects a set of evidence documents $\mathcal{D}_q$ for a query $q$, and a generator produces a multi-paragraph response grounded in these documents:
\[
y = \mathcal{G}(q, \mathcal{D}_q).
\]
For each document $d \in \mathcal{D}_q$, we decompose it into a set of atomic claims
\[
\mathcal{C}_d = \{c_1, c_2, \ldots, c_{n_d}\},
\]
where each $c_i$ represents a semantically irreducible and independently verifiable atomic claim.
The union of extracted atomic claims forms the initial claim pool for subsequent structured evidence modeling.

To handle noisy, redundant, and potentially conflicting evidence in long-form RAG, we propose \textbf{ArbGraph}, a framework that transforms the conventional linear RAG pipeline into an \emph{arbitrate-first, generate-later} paradigm.
Instead of implicitly integrating potentially conflicting evidence during generation, ArbGraph explicitly performs evidence-level arbitration prior to decoding.
This design separates \emph{credibility reasoning} from \emph{text generation}, enabling more robust handling of noisy and contradictory evidence.

\begin{figure*}[t]
    \centering
    \includegraphics[width=\linewidth]{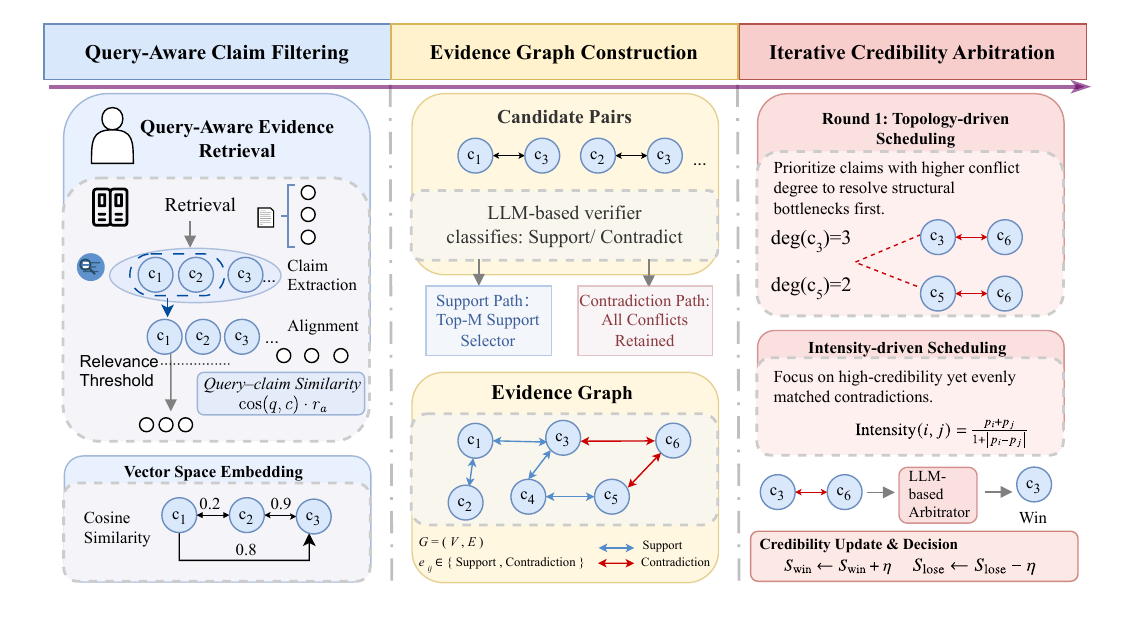}
    \caption{Overview of the ArbGraph framework. The pipeline comprises three stages: query-aware claim filtering, evidence graph construction, and iterative credibility arbitration, which together realize an arbitrate-first, generate-later paradigm.}
    \label{fig:framework}
\end{figure*}

As illustrated in Figure~\ref{fig:framework}, ArbGraph consists of three stages:
(1) query-aware claim filtering,
(2) evidence graph construction, and
(3) iterative credibility arbitration.
These stages progressively refine the evidence pool and produce a validated claim set $\mathcal{V}_{\text{val}}$, which serves as the trusted context for long-form generation.
We briefly outline each stage below and defer implementation details to subsequent sections.

\paragraph{Stage I: Query-Aware Claim Filtering.}
ArbGraph first converts retrieved documents into atomic claims and filters them with respect to the query.
This step removes background or weakly related information, ensuring that subsequent reasoning operates on a focused and task-relevant evidence set.
By controlling the scale and relevance of claims early on, it also improves both efficiency and reliability of later stages.

\paragraph{Stage II: Evidence Graph Construction.}
The filtered claims are organized into an \emph{Evidence Graph} that captures logical relationships between claims.
In particular, the graph encodes both supportive and contradictory interactions, providing an explicit structural view of evidence dependencies.
This representation enables reasoning beyond independent claim scoring by exposing how claims reinforce or conflict with each other.

\paragraph{Stage III: Iterative Credibility Arbitration.}
Given the constructed graph, ArbGraph performs iterative arbitration over conflicting claims to refine their credibility.
The arbitration process resolves high-impact conflicts using contextual evidence and progressively suppresses less reliable claims.
The remaining high-credibility claims form the validated set $\mathcal{V}_{\text{val}}$, which is then used as grounded input for final generation.

\subsection{Atomic Claim Extraction and Semantic Alignment}

The efficacy of evidence-level arbitration hinges on the precision of the underlying reasoning units. Reasoning directly over coarse-grained passages often introduces linguistic noise and obscures fine-grained factual dependencies, making it difficult to isolate specific points of contention. \textbf{ArbGraph} addresses this by decomposing evidence documents into atomic claims---the fundamental, irreducible units of information that facilitate explicit comparison and conflict resolution. This decomposition is particularly vital because factual contradictions are typically localized within specific attributes, such as temporal markers, geospatial coordinates, or quantitative values, which are frequently embedded within dense, non-atomic textual contexts. This step establishes the basis for query-aware claim filtering, while the final relevance-based selection is performed later during node selection.

\paragraph{Atomic Decomposition.}
Given a set of retrieved documents $\mathcal{D}_q$ for a query $q$, we employ an instruction-tuned Large Language Model (LLM) to transform each document $d \in \mathcal{D}_q$ into a set of atomic claims $\mathcal{C}_d = \{ c_1, \ldots, c_{n_d} \}$. Each claim is formulated as a self-contained assertion expressing a single, independently verifiable fact. For instance, a sentence such as \textit{``Tesla was founded in 2003 and is headquartered in Austin''} is bifurcated into two discrete units: one specifying the founding year and another identifying the corporate headquarters.

The granularity of this decomposition is a critical determinant of the resulting evidence graph's topology. We recognize a fundamental trade-off: over-segmentation may fragment semantically coherent statements into excessively sparse units, leading to a disconnected graph where relation verification becomes numerically unstable; conversely, under-segmentation tends to entangle multiple factual dimensions within a single node, thereby masking potential contradictions and diluting the precision of the arbitration process. Consequently, ArbGraph is designed to produce \textit{minimal yet self-contained} factual units that preserve semantic integrity while maximizing the resolution of conflict detection. While LLM-based extraction may introduce occasional heuristic errors, the systemic impact of such noise is attenuated in subsequent stages through multi-source aggregation, which prioritizes consistent evidence patterns over idiosyncratic segmentation noise.

\paragraph{Semantic Alignment and Normalization.}
Information gathered from diverse sources often exhibits significant lexical variance despite expressing identical propositions. To mitigate this redundancy, we perform semantic normalization by mapping claims into a latent vector space using a sentence encoder $f(\cdot)$. Claims that exceed a conservative similarity threshold and maintain consistency across key entities and attributes are merged into a unified canonical representation.

This normalization step is not merely a data compression technique but a prerequisite for valid evidence weighting. Without it, the frequency of a claim's appearance---often a byproduct of document popularity or paraphrasing---could be misinterpreted as independent corroboration. By consolidating these duplicates, we ensure that the perceived strength of a claim reflects a genuine consensus across distinct information providers rather than the mere repetition of a single source's phrasing.

In the resulting evidence graph, each canonical cluster is instantiated as a single \textbf{claim node} that aggregates all source attributions, while claims representing genuine factual disagreements are intentionally preserved as distinct, competing nodes. This architecture explicitly decouples redundancy from disagreement: support edges represent independent cross-source verification, whereas contradiction edges isolate fundamental inconsistencies. This structural refinement enhances the density of meaningful interactions within the graph, providing a stable foundation for the iterative credibility arbitration that follows.

\subsection{Evidence Graph Construction}

Given the normalized and query-filtered claim nodes, we construct an \emph{Evidence Graph}
\[
\mathcal{G}_E = (\mathcal{V}, \mathcal{E}),
\]
where $\mathcal{V}$ denotes canonical claim nodes and $\mathcal{E}$ represents verified logical relations between them.
This graph-based formulation enables explicit modeling of inter-claim dependencies,
allowing downstream arbitration to operate over structured evidence rather than independent claims.

Compared to flat aggregation or independent scoring approaches,
the graph representation explicitly captures how claims interact,
making both agreement and conflict observable at the structural level.
This is particularly important in long-form RAG settings,
where evidence may be redundant, noisy, or mutually inconsistent.

We consider two types of relations: \emph{Support}, indicating mutual reinforcement,
and \emph{Contradiction}, capturing logical conflicts.
The edge set is thus partitioned as $\mathcal{E}=\mathcal{E}_{\text{sup}} \cup \mathcal{E}_{\text{con}}$.

The construction process consists of candidate pair mining and relation verification.

\subsubsection{Query-Aware Node Selection}

The node set $\mathcal{V}$ is inherited from Stage I, where query-aware filtering ensures that only task-relevant claims are retained.
Specifically, we map the query $q$ and each claim $c_i$ into a shared embedding space via a sentence encoder $f(\cdot)$,
and retain a claim as a node if its semantic similarity to the query exceeds a predefined threshold:
\begin{equation}
\cos\bigl(f(q), f(c_i)\bigr) \geq \tau_q,
\label{eq:query-aware-filter}
\end{equation}
where $\tau_q$ denotes the query relevance threshold.

This step effectively bounds the graph size and reduces the likelihood of introducing spurious relations during edge construction.
In addition to efficiency considerations, it also improves the \emph{signal-to-noise ratio} of the graph.
Including irrelevant or weakly related claims would introduce noisy edges that may distort the structural patterns of support and contradiction,
thereby degrading the reliability of subsequent arbitration.

\subsubsection{Threshold-Based Candidate Pair Mining}
Naïve pairwise verification incurs quadratic complexity ($O(|\mathcal{V}|^2)$),
which is impractical for large claim sets.
To address this, we adopt a similarity-based pruning strategy:
\begin{equation}
\mathcal{P}_{\text{cand}} = \{ (c_i, c_j) \mid \cos(f(c_i), f(c_j)) \geq \tau_{\text{sim}} \},
\label{eq:candidate-mining}
\end{equation}
where $\tau_{\text{sim}}$ controls the semantic proximity required for further verification.

This design is motivated by the observation that meaningful logical interactions—whether supportive or contradictory—
typically arise among semantically related claims that refer to the same entities or attributes.
In contrast, semantically distant claims are unlikely to exhibit direct logical relationships.

By restricting verification to local semantic neighborhoods,
we significantly reduce computational cost while preserving informative interactions.
This also implicitly enforces a form of \emph{structural sparsity},
which is beneficial for downstream reasoning by preventing the graph from becoming overly dense and noisy.

At the same time, $\tau_{\text{sim}}$ governs a trade-off between coverage and reliability in graph construction.
An overly dense graph (i.e., low $\tau_{\text{sim}}$) may introduce spurious interactions and amplify noise propagation during arbitration,
while an overly sparse graph (i.e., high $\tau_{\text{sim}}$) may miss critical relationships between claims,
reducing the effectiveness of conflict detection.
By tuning $\tau_{\text{sim}}$, ArbGraph ensures that arbitration operates over informative yet well-controlled semantic neighborhoods.

\subsubsection{Relation Verification and Edge Instantiation}
For each candidate pair $(c_i, c_j) \in \mathcal{P}_{\text{cand}}$,
a verifier model predicts their relation as \emph{Support}, \emph{Contradiction}, or \emph{Neutral},
along with an associated confidence score.

To ensure reliability, only relations with confidence exceeding a threshold $\tau_{\text{conf}}$
are instantiated as edges in $\mathcal{E}$.
This conservative filtering mitigates noise from imperfect classification,
especially in cases involving numerical or temporal reasoning.

Beyond filtering, the graph structure itself provides robustness through \emph{multi-source aggregation}.
Consistent relations supported by multiple independent claim pairs are reinforced through redundant connections,
while isolated or spurious edges are less likely to influence global reasoning.
To further control graph density, we retain at most $M$ support edges for each node, preventing excessive propagation over weak or noisy interactions.

Furthermore, explicitly separating support and contradiction edges enables downstream arbitration
to reason over agreement and conflict in a structured manner,
rather than relying on implicit signal aggregation within text.
This separation is critical for accurately resolving inconsistencies in complex evidence settings.

\subsection{Iterative Credibility Arbitration}
\label{sec:arbitration}

Given the evidence graph $\mathcal{G}_E$, we perform iterative credibility arbitration to obtain a high-confidence subset of claims.
Each claim $c_i \in \mathcal{V}$ is associated with a credibility score $p_i \in [0,1]$.
For numerical stability, we parameterize credibility in the logit space:
\begin{equation}
s_i = \text{logit}(p_i) = \log \frac{p_i}{1-p_i}.
\label{eq:logit-credibility}
\end{equation}

At round $t$, logits are converted to probabilities via $p_i^{(t)} = \sigma(s_i^{(t)})$,
which are used for both arbitration scheduling and validation.
The iterative design reflects the interdependent nature of contradictions,
where resolving one conflict may affect the credibility of others.

The full procedure is summarized in Algorithm~\ref{alg:refinement}.

\begin{algorithm}[t]
\small
\caption{ArbGraph: Iterative Credibility Arbitration}
\label{alg:refinement}
\begin{algorithmic}[1]
\Require Evidence graph $\mathcal{G}_E = (\mathcal{V}, \mathcal{E}_{\text{sup}} \cup \mathcal{E}_{\text{con}})$, step size $\eta$, threshold $\tau_{\text{accept}}$, budget $k$
\Ensure Validated claim set $\mathcal{V}_{\text{val}}$

\State \textbf{Initialize:} $s_i \gets \text{logit}(p_i^{(0)})$ for all $c_i \in \mathcal{V}$

\For{$t = 1$ \textbf{to} $T$}
    \State $p_i \gets \sigma(s_i)$ for all $c_i \in \mathcal{V}$ \Comment{Current credibility scores}
    \State $\mathcal{P}_{\text{active}} \gets \emptyset$
    
    \For{$(c_i, c_j) \in \mathcal{E}_{\text{con}}$ \textbf{with} $p_i, p_j > \tau_{\text{accept}}$}
        \State $I_{ij} \gets \dfrac{p_i + p_j}{1 + |p_i - p_j|}$
        \State Add $(c_i, c_j, I_{ij})$ to $\mathcal{P}_{\text{active}}$
    \EndFor
    
    \State Select top-$k$ conflicting pairs $\mathcal{P}_{\text{top}} \subset \mathcal{P}_{\text{active}}$ based on $I_{ij}$
    
    \For{$(c_i, c_j) \in \mathcal{P}_{\text{top}}$}
        \State $\mathcal{E}_{\text{ctx}} \gets \mathcal{N}_{\text{sup}}(c_i) \cup \mathcal{N}_{\text{sup}}(c_j)$
        \State $(w, c_{\text{win}}, c_{\text{lose}}) \gets \textsc{Arbitrate}(c_i, c_j, \mathcal{E}_{\text{ctx}})$
        \If{$w = 1$}
            \State $s_{\text{win}} \gets s_{\text{win}} + \eta;\quad s_{\text{lose}} \gets s_{\text{lose}} - \eta$
        \EndIf
    \EndFor
\EndFor

\State \Return $\mathcal{V}_{\text{val}} \gets \{ c_i \mid \sigma(s_i) \geq \tau_{\text{accept}} \}$
\end{algorithmic}
\end{algorithm}

\paragraph{Active Contradiction Mining.}
At each round, we prioritize conflicts among currently plausible claims.
For each contradiction edge $(c_i, c_j) \in \mathcal{E}_{\text{con}}$, we compute an intensity score
\begin{equation}
I_{ij} = \frac{p_i + p_j}{1 + |p_i - p_j|},
\label{eq:intensity-score}
\end{equation}
which favors pairs with both high credibility and small confidence gaps.

This prioritization focuses computation on high-impact ambiguities,
while avoiding unnecessary updates on low-confidence or easily separable cases.

\paragraph{Gated Arbitration with Supporting Context.}
For each selected pair, we perform context-aware arbitration using local graph structure.
Specifically, we construct a supporting context
\[
\mathcal{E}_{\text{ctx}} = \mathcal{N}_{\text{sup}}(c_i) \cup \mathcal{N}_{\text{sup}}(c_j),
\]
where $\mathcal{N}_{\text{sup}}(c)$ denotes support neighbors of $c$.

An LLM-based arbitrator evaluates $(c_i, c_j)$ conditioned on $\mathcal{E}_{\text{ctx}}$
and outputs a winner $c_{\text{win}}$, a loser $c_{\text{lose}}$, and a binary gate $w \in \{0,1\}$.
The gate indicates whether sufficient evidence is available to support a confident decision.
This mechanism prevents premature or unreliable updates when the supporting context is insufficient,
thereby reducing the risk of propagating incorrect decisions across iterations.
In practice, the gating decision is implemented using a heuristic confidence threshold,
which is empirically stable within a reasonable range.

When $w=1$, we apply symmetric logit updates:
\[
s_{\text{win}} \gets s_{\text{win}} + \eta, \qquad
s_{\text{lose}} \gets s_{\text{lose}} - \eta .
\]

This update rule ensures stable and bounded adjustments of credibility scores,
while preserving relative ordering without introducing bias drift.
Together with the gating mechanism, it enables ArbGraph to perform reliable credibility updates
driven by sufficiently supported evidence rather than local or noisy signals.

\paragraph{Stability Mechanisms.}
To mitigate noise in individual arbitration decisions, ArbGraph incorporates several stabilizing components:
(1) gated updates that suppress low-confidence decisions;
(2) top-$k$ scheduling that limits the influence of spurious conflicts;
(3) iterative refinement that allows decisions to be revised as new evidence is incorporated;
and (4) contextual arbitration grounded in multi-source supporting evidence.

\paragraph{Validated Claim Set.}
After $T$ rounds, we obtain the validated claim set
\begin{equation}
\mathcal{V}_{\text{val}} = \{ c_i \mid \sigma(s_i^{(T)}) \geq \tau_{\text{accept}} \},
\label{eq:validated-set}
\end{equation}
which serves as the conflict-aware evidence context for final generation.

\subsection{Computational Efficiency Discussion}
\label{sec:efficiency}

Pre-generation evidence arbitration inevitably introduces additional computation beyond a standard retrieve-and-generate pipeline. However, ArbGraph is designed so that this overhead remains explicitly bounded by structural constraints at each stage, rather than growing through open-ended reasoning or repeated retrieval--generation loops.

First, query-aware claim filtering limits the size of the initial claim pool by removing irrelevant or weakly related information before graph construction. Second, relation verification is restricted to local semantic neighborhoods through threshold-based candidate mining parameterized by $\tau_{\text{sim}}$, avoiding exhaustive pairwise comparison over all claims. In addition, retaining at most $M$ support edges bounds graph density and prevents excessive propagation over noisy or weak interactions. Third, during arbitration, LLM-based decisions are allocated selectively rather than exhaustively: only the top-$k$ high-impact conflicts are considered in each round, and confidence-based gating suppresses unnecessary updates on marginal or weakly supported cases. Together, these mechanisms constrain the effective search space for arbitration and reduce the number of expensive verification and decision steps.

Under this design, the computational cost of ArbGraph is governed primarily by the size of the filtered claim graph and by the arbitration budget, rather than by unbounded search trajectories. Compared with agentic baselines such as ReAct or MCTS-style methods, which may involve multiple rounds of retrieval or long-horizon reasoning paths, ArbGraph operates under a fixed retrieval budget and a bounded number of verification and arbitration steps. We therefore view its cost profile as more controlled and predictable, especially in settings where the main challenge lies in resolving conflicting evidence rather than repeatedly expanding the search process.

Overall, the inference cost of ArbGraph scales with the amount of retained evidence and the complexity of evidence conflicts, while remaining explicitly constrained by filtering, sparsification, and bounded arbitration. This makes ArbGraph a practically manageable reliability mechanism for long-form RAG, rather than an open-ended reasoning procedure.
\section{Experiments}
This section presents a systematic empirical evaluation of ArbGraph for long-form retrieval-augmented generation. 
We evaluate its performance from multiple perspectives, including overall generation quality, robustness under noisy retrieval, the contribution of key components, and sensitivity to graph construction and arbitration parameters. 
We begin by introducing the experimental settings, followed by the main results, ablation studies, and detailed analyses on LongFact and RAGChecker.

\subsection{Experimental Settings}
\label{sec:exp-settings}

\subsubsection{Datasets}
\label{sec:datasets}

We evaluate ArbGraph on two complementary benchmarks, \textbf{LongFact}~\cite{wei2024longform} and \textbf{RAGChecker}~\cite{ru2024ragchecker}, enabling a comprehensive assessment of factuality, grounding, and robustness in long-form retrieval-augmented generation.

\textbf{LongFact} is a manually curated benchmark for long-form factual generation, containing complex information-seeking queries spanning 38 knowledge domains. Following prior practice, we group these domains into eight broader categories for analysis. A key advantage of LongFact is that each reference answer is annotated with \emph{atomic information units}, which enables fine-grained evaluation of factual coverage and evidence completeness at the level of atomic claims. This benchmark is therefore particularly suitable for assessing whether ArbGraph improves factual recall and preserves informative content in long-form responses.

\textbf{RAGChecker} provides a complementary diagnostic benchmark for retrieval-augmented generation. It covers queries from 10 public domains and evaluates systems from multiple perspectives, including grounding quality, retrieval usage, and robustness to noisy evidence. Since the original benchmark is primarily designed for short-form responses, we follow its diagnostic protocol while extending it to long-form generation to better match our task formulation. This benchmark is especially useful for analyzing whether ArbGraph improves faithfulness and reduces the impact of both relevant and irrelevant retrieval noise.

Taken together, LongFact emphasizes \emph{fine-grained factual coverage} in long-form answers, whereas RAGChecker provides a \emph{diagnostic view of grounding robustness}. Their combination enables a comprehensive evaluation of ArbGraph from both generation quality and evidence reliability perspectives.

\subsubsection{Evaluation Metrics}
\label{sec:metrics}

We evaluate model performance using a suite of metrics covering three aspects: generative informativeness, factual grounding, and robustness to retrieval noise.

For generative informativeness, we report Fact Recall (FR) and Information Density (ID). Following long-form factual generation evaluation~\cite{fan2019eli5}, FR measures the proportion of ground-truth atomic facts covered by the generated answer, reflecting factual completeness. ID measures factual efficiency, defined as the number of atomic facts per token.

For factual grounding, we adopt the diagnostic metrics from RAGChecker. Faithfulness measures the proportion of generated claims that are supported by the retrieved evidence, indicating how reliably the model grounds its output in external context. Context Utilization measures how well the retrieved evidence captures the ground-truth facts that should appear in the response, reflecting the effectiveness of retrieval usage.

For robustness to noisy retrieval, we report two noise-sensitivity metrics from RAGChecker: Relevant Noise Sensitivity (Noise-S) and Irrelevant Noise Sensitivity (Noise-I). Noise-S evaluates how susceptible the model is to misleading but topically related evidence, while Noise-I measures the impact of distracting evidence that is unrelated to the query. Lower sensitivity indicates stronger robustness under noisy retrieval conditions.

Furthermore, to analyze knowledge attribution, we report Hallucination and Self-knowledge. Hallucination quantifies the proportion of generated claims that are unsupported or incorrect with respect to the retrieved evidence. Self-knowledge measures correct claims produced from the model's internal parametric knowledge rather than from the retrieved context.


\subsubsection{Baseline Models}
\label{sec:baselines}

To evaluate the impact of explicit evidence-level arbitration under long-form generation settings, we compare ArbGraph with representative baselines that incorporate reasoning or verification to handle potentially conflicting evidence. 
Recent graph-based RAG approaches (e.g., GraphRAG~\cite{edge2024graphrag}, HippoRAG~\cite{jimenezgutierrez2024hipporag}) have shown strong performance in enhancing retrieval coverage and multi-hop connectivity. However, they primarily focus on improving information organization and connectivity, without explicitly addressing evidence-level conflict resolution. 
In addition, methods such as Self-RAG~\cite{asai2023self} introduce token-level verification during generation, but are mainly designed for short-form QA settings. 
Therefore, we compare ArbGraph against representative methods for long-form RAG under similar evaluation settings. The selected baselines span three categories: Naive RAG, Reasoning RAG, and Agentic RAG, reflecting a progression from static pipelines to more autonomous reasoning frameworks.

(1) \textbf{Naive RAG}.  
Standard RAG~\cite{lewis2020rag} serves as a basic baseline, following a retrieve-and-generate pipeline without intermediate reasoning or verification. It directly concatenates retrieved documents into the context window, making the generation process susceptible to input noise and implicit factual conflicts.

(2) \textbf{Reasoning RAG}.  
This category introduces explicit reasoning steps to improve generation quality.  
CoT (Chain-of-Thought)~\cite{wei2022chain} prompts the model to produce intermediate reasoning trajectories prior to the final answer, encouraging deeper parametric reasoning over the provided context.  
ReAct~\cite{yao2022react} interleaves reasoning with active retrieval, enabling iterative refinement of context. However, both methods rely on the language model's implicit ability to resolve inconsistencies among retrieved evidence during generation.

(3) \textbf{Agentic RAG}.  
These methods incorporate agent-like behaviors, such as self-verification and iterative refinement, to handle complex queries.  
CRITIC~\cite{gou2023critic} adopts a generate--verify--correct loop with external tools, attempting to rectify factual errors after the initial generation phase.  
FIRE~\cite{xie2025fire} improves factuality through fine-grained, sequential verification during generation, primarily emphasizing compact and controlled outputs rather than explicit, structure-aware evidence consolidation.  
In our implementation, we follow its original formulation by prompting the model to perform step-by-step verification during generation, rather than applying it as a post-hoc filtering mechanism on top of naive RAG outputs.  
MCTS-RAG~\cite{hu2025mctsr} formulates long-form generation as a principled search process, using Monte Carlo Tree Search to explore alternative reasoning trajectories and evaluate state values.

\subsubsection{Implementation Details}
\label{sec:impl}

We adopt two instruction-tuned large language models as generators:
Qwen3-4B-Instruct~\cite{yang2025qwen3} and LLaMA-3.1-8B-Instruct~\cite{grattafiori2024llama3}.
Unless otherwise specified, all components of ArbGraph, including claim extraction, relation verification, and arbitration, share the same configuration across both backbones.

To ensure a fair comparison, we use a unified prompting strategy and decoding setup for all models.
Specifically, we employ deterministic decoding with temperature set to 0, and fix the maximum generation length for long-form responses.
For retrieval, we use a fixed Wikipedia-based corpus with a top-$k=5$ setting for all methods and retrieve relevant documents by the official Wikipedia API~\footnote{\url{https://www.mediawiki.org/wiki/API:Main_page}}.
Given that both backbones exhibit consistent relative performance trends in our main experiments, we conduct ablation studies and detailed analyses on Qwen3-4B-Instruct for efficiency, while reporting the main results on both models.

Atomic claims are embedded using \texttt{bge-large-en-v1.5}, and cosine similarity is employed for both query-aware filtering and claim matching.
We set the query relevance threshold to $\tau_q=0.3$ and adopt a threshold-based candidate retrieval strategy for relation mining.
All contradiction edges verified by the LLM are preserved, while support edges are globally pruned by retaining only the top-$M$ edges ranked by semantic similarity (with $M=60$ in all experiments).
The iterative credibility arbitration module runs with an arbitration budget of $k=3$ per round, step size $\eta=0.8$, and acceptance threshold $\tau_{\text{accept}}=0.3$.

\begin{table*}[t]
\centering
\small 

\caption{Main results on LongFact across eight domains using two backbones. We report (a) Fact Recall (FR) and (b) Information Density (ID). Sci.: Science, Tech.: Technology, Med.: Medicine, Cult.: Culture, Comm.: Communication, Life.: Lifestyle. The best results in each backbone group are highlighted in bold.}
\begin{tabular}{@{} l *{9}{c} @{}}
\toprule
\textbf{Method} & \textbf{Sci.} & \textbf{Tech.} & \textbf{Med.} & \textbf{Law} & \textbf{Cult.} & \textbf{Event} & \textbf{Comm.} & \textbf{Life.} & \textbf{Avg.} \\
\midrule

\multicolumn{10}{c}{\textit{(a) Fact Recall (FR) $\uparrow$}} \\
\midrule
\multicolumn{10}{c}{\textit{Backbone: Qwen3-4B}} \\
Standard RAG & 64.9 & 73.7 & 64.7 & 61.6 & 58.5 & 56.7 & 56.6 & 68.6 & 63.5 \\
ReAct        & 65.6 & 59.2 & 71.3 & 62.4 & 64.4 & 66.0 & 50.6 & 70.8 & 63.3 \\
CoT          & 65.7 & 70.0 & 65.6 & 62.5 & 55.2 & 58.5 & 55.8 & 69.4 & 63.1 \\
CRITIC       & 78.4 & 79.7 & 82.4 & 73.3 & 76.6 & 75.2 & 69.3 & 75.4 & 76.3 \\
FIRE         & 79.7 & 76.6 & 82.8 & 77.1 & 80.2 & 79.1 & 74.0 & 81.2 & 78.8 \\
MCTS-RAG     & 80.4 & 77.4 & 85.2 & 75.8 & 82.4 & 79.2 & 76.3 & 77.9 & 79.2 \\
\textbf{Ours} & \textbf{83.5} & \textbf{84.4} & \textbf{86.2} & \textbf{83.3} & \textbf{82.5} & \textbf{79.8} & \textbf{84.7} & \textbf{81.6} & \textbf{83.3} \\

\addlinespace[0.2em]
\multicolumn{10}{c}{\textit{Backbone: LLaMA-3.1-8B}} \\
Standard RAG & 59.6 & 68.8 & 69.8 & 66.0 & 62.2 & 59.5 & 57.7 & 72.1 & 64.6 \\
CoT          & 69.1 & 56.1 & 68.3 & 61.9 & 66.6 & 69.0 & 73.4 & 60.8 & 64.7 \\
ReAct        & 69.5 & 69.2 & 71.9 & 68.7 & 67.3 & 68.3 & 66.8 & 70.4 & 69.0 \\
CRITIC       & 78.2 & 69.2 & 78.1 & 71.3 & 70.1 & 74.3 & 69.2 & 75.2 & 73.3 \\
FIRE         & 79.2 & 78.4 & 80.5 & 77.1 & 76.2 & 75.1 & 78.2 & 73.6 & 78.3 \\
MCTS-RAG     & 83.8 & 81.6 & 84.0 & 82.5 & 80.5 & 81.9 & 77.1 & 81.9 & 81.7 \\
\textbf{Ours} & \textbf{84.2} & \textbf{85.1} & \textbf{86.1} & \textbf{85.3} & \textbf{84.0} & \textbf{83.0} & \textbf{85.9} & \textbf{84.7} & \textbf{84.9} \\

\midrule
\multicolumn{10}{c}{\textit{(b) Information Density (ID) $\uparrow$}} \\
\midrule
\multicolumn{10}{c}{\textit{Backbone: Qwen3-4B}} \\
Standard RAG & 79.1  & 99.2  & 83.6  & 77.3  & 77.6  & 72.5  & 73.2  & 86.2  & 81.4 \\
ReAct        & 72.1  & 70.3  & 81.1  & 62.0  & 70.5  & 56.9  & 60.2  & 86.4  & 70.6 \\
CoT          & 100.1 & 115.9 & 111.7 & 91.3  & 93.9  & 95.5  & 96.6  & 124.0 & 104.3 \\
CRITIC       & 71.1  & 78.5  & 81.0  & 70.6  & 71.4  & 76.1  & 71.9  & 83.8  & 75.5 \\
FIRE         & 229.9 & 232.5 & 107.6 & 240.9 & 155.1 & 107.8 & 93.7  & \textbf{219.7} & 177.2 \\
MCTS-RAG     & 97.8  & 120.0 & 95.6  & 76.6  & 92.8  & 98.0  & 93.8  & 85.8  & 95.2 \\
\textbf{Ours} & \textbf{247.3} & \textbf{235.5} & \textbf{140.6} & \textbf{241.2} & \textbf{156.4} & \textbf{132.5} & \textbf{124.5} & 190.2 & \textbf{187.6} \\

\addlinespace[0.2em]
\multicolumn{10}{c}{\textit{Backbone: LLaMA-3.1-8B}} \\
Standard RAG & 42.7  & 51.3  & 54.3  & 47.6  & 43.4  & 38.0  & 36.3  & 58.5  & 46.9 \\
CoT          & 56.9  & 61.0  & 77.7  & 57.8  & 54.8  & 51.7  & 48.7  & 53.5  & 57.8 \\
ReAct        & 70.4  & 66.6  & 76.2  & 61.8  & 72.0  & 48.1  & 63.0  & 73.5  & 67.1 \\
CRITIC       & 79.0  & 68.7  & 85.7  & 63.9  & 77.6  & 52.5  & 64.2  & 79.1  & 72.0 \\
FIRE         & 162.4 & 170.1 & 158.9 & 175.2 & 160.2 & 161.6 & 155.1 & 158.1 & 165.7 \\
MCTS-RAG     & 61.3  & 61.9  & 62.0  & 59.5  & 59.5  & 58.3  & 54.6  & 62.9  & 60.0 \\
\textbf{Ours} & \textbf{178.6} & \textbf{191.0} & \textbf{176.2} & \textbf{195.7} & \textbf{174.1} & \textbf{177.5} & \textbf{171.2} & \textbf{176.0} & \textbf{180.0} \\

\bottomrule
\end{tabular}

\label{tab:main_results}
\end{table*}

\begin{table*}[t]
\centering
\small
\setlength{\tabcolsep}{3.5pt} 

\caption{Overall performance comparison across paradigms and backbones. FR: Fact Recall; ID: Information Density; Util.: Evidence Utilization; Noise-S/I: Support/Irrelevant Noise; Hallu.: Hallucination Rate; Self-K: Self-Knowledge; Faith.: Faithfulness.}
\begin{tabular}{@{} l l cccccccc @{}}
\toprule
\textbf{Paradigm} & \textbf{Model} 
& \textbf{FR}$\uparrow$ 
& \textbf{ID}$\uparrow$ 
& \textbf{Util.}$\uparrow$ 
& \textbf{Noise-S}$\downarrow$ 
& \textbf{Noise-I}$\downarrow$ 
& \textbf{Hallu.}$\downarrow$ 
& \textbf{Self-K}$\downarrow$ 
& \textbf{Faith.}$\uparrow$ \\
\midrule

\multicolumn{10}{c}{\textit{Backbone: Qwen3-4B}} \\
\midrule
\textbf{Naive} 
& Standard RAG 
& 54.3 & 75.8 & 31.5 & 7.9 & 7.7 & 51.4 & 6.6 & 32.2 \\
\addlinespace
\multirow{2}{*}{\textbf{Reasoning}} 
& ReAct 
& 54.4 & 45.0 & 28.9 & 7.1 & 9.4 & 55.2 & 6.5 & 35.1 \\
& CoT 
& 57.6 & 110.6 & 28.4 & 7.0 & 7.9 & 49.5 & 6.8 & 30.1 \\
\addlinespace
\multirow{3}{*}{\textbf{Agentic}} 
& CRITIC 
& 54.8 & 41.8 & 25.0 & 4.8 & 5.2 & 45.0 & 6.9 & 35.0 \\
& FIRE 
& 70.7 & 59.6 & 34.6 & 11.6 & 8.1 & 48.6 & 7.1 & 37.8 \\
& MCTS-RAG 
& 72.0 & 98.7 & 32.6 & 7.0 & 10.4 & 36.9 & 4.6 & 50.5 \\
\addlinespace
\textbf{Ours} 
& \textbf{ArbGraph} 
& \textbf{74.5} & \textbf{135.4} & \textbf{36.6} & \textbf{4.2} & \textbf{4.9} & \textbf{33.3} & \textbf{4.4} & \textbf{53.8} \\

\midrule
\multicolumn{10}{c}{\textit{Backbone: LLaMA-3.1-8B}} \\
\midrule

\textbf{Naive} 
& Standard RAG 
& 54.7 & 51.7 & 29.7 & 12.5 & 6.0 & 59.9 & 6.9 & 29.2 \\
\addlinespace
\multirow{2}{*}{\textbf{Reasoning}} 
& CoT 
& 59.8 & 57.8 & 30.4 & 9.4 & 6.2 & 52.5 & 6.8 & 29.8 \\
& ReAct 
& 62.1 & 62.6 & 31.5 & 12.2 & 7.7 & 59.4 & 7.1 & 25.8 \\
\addlinespace
\multirow{3}{*}{\textbf{Agentic}} 
& CRITIC 
& 62.5 & 78.8 & 32.9 & 13.6 & 5.2 & 56.1 & 7.0 & 31.0 \\
& MCTS-RAG 
& 71.3 & 96.6 & 33.5 & 9.4 & 9.8 & 43.1 & 5.9 & 39.7 \\
& FIRE 
& 68.7 & 88.2 & 33.1 & 8.7 & 4.7 & 41.8 & 5.3 & 33.6 \\
\addlinespace
\textbf{Ours} 
& \textbf{ArbGraph} 
& \textbf{75.6} & \textbf{99.3} & \textbf{41.2} & \textbf{4.2} & \textbf{3.6} & \textbf{36.3} & \textbf{4.6} & \textbf{51.5} \\
\bottomrule
\end{tabular}

\label{tab:overall-performance}
\end{table*}
\subsection{Main Results}
\label{sec:main-results}

ArbGraph consistently outperforms all baselines across both benchmarks and backbones, achieving improvements in factual coverage, information density, and robustness under noisy retrieval. Detailed results are reported in Table~\ref{tab:main_results} and Table~\ref{tab:overall-performance}.

\textbf{Strong performance in long-form generation.}
Across both benchmarks, ArbGraph achieves the best overall performance under both backbones. On LongFact, prior methods exhibit a trade-off between factual coverage and generation efficiency, where higher recall often comes with less compact outputs, while density-oriented methods tend to miss supporting facts. Unlike these approaches, ArbGraph improves both Fact Recall (FR) and Information Density (ID) simultaneously. Under Qwen3-4B, it increases FR from 79.2 to 83.3 (vs. MCTS-RAG) and ID from 177.2 to 187.6 (vs. FIRE), with similar gains observed under LLaMA-3.1-8B. Notably, these improvements over the strongest baselines are statistically significant ($p < 0.01$). 
This joint improvement can be attributed to ArbGraph’s ability to filter out redundant or unreliable claims while preserving complementary evidence through structured arbitration.
As a result, the model can generate more compact responses without sacrificing factual completeness.

\textbf{Consistency across backbones.}
ArbGraph shows consistent improvements on both Qwen3-4B and LLaMA-3.1-8B, indicating that the gains are largely model-agnostic and stem from improved evidence consolidation rather than model scale.

\textbf{Improved trustworthiness.}
On RAGChecker, ArbGraph achieves the highest Faithfulness and lowest Hallucination across both backbones, while maintaining strong factual recall. It also attains the best evidence utilization, suggesting more effective consolidation of useful evidence. This result indicates that explicit evidence-level arbitration helps filter unreliable or conflicting evidence before generation.

\textbf{Robustness to noise.}
ArbGraph demonstrates strong robustness under noisy retrieval, achieving the lowest noise sensitivity across both backbones. For example, under Qwen3-4B, MCTS-RAG obtains FR 72.0 but suffers from high irrelevant-noise sensitivity (10.4), whereas ArbGraph achieves higher FR (74.5) with substantially lower sensitivity (4.9). This suggests that resolving conflicts prior to generation reduces the impact of both relevant and irrelevant noise.

Overall, these results demonstrate that resolving conflicts at the evidence level provides a more reliable foundation for long-form generation than generation-time heuristics.

\begin{table*}[t]
\centering
\small

\caption{Ablation results on the LongFact dataset. We show the impact of different components on (a) Fact Recall and (b) Information Density across eight domains. Bold numbers indicate the best performance.}
\begin{tabular}{@{} l *{9}{c} @{}}
\toprule
\textbf{Method} & \textbf{Sci.} & \textbf{Tech.} & \textbf{Med.} & \textbf{Law} & \textbf{Cult.} & \textbf{Event} & \textbf{Comm.} & \textbf{Life.} & \textbf{Avg.} \\
\midrule
\multicolumn{10}{c}{\textit{(a) Fact Recall (FR) $\uparrow$}} \\
\midrule
\textbf{ArbGraph (Ours)} & \textbf{83.5} & \textbf{84.4} & \textbf{86.2} & \textbf{83.3} & \textbf{82.5} & \textbf{79.8} & \textbf{84.7} & \textbf{81.6} & \textbf{83.3} \\
w/o Evidence Graph       & 65.6 & 68.2 & 64.8 & 63.4 & 66.1 & 67.5 & 58.9 & 72.4 & 66.1 \\
w/o Soft Arbitration    & 77.1 & 74.4 & 78.0 & 65.4 & 74.2 & 64.1 & 66.9 & 73.2 & 72.2 \\
w/o Initial Screening   & 73.0 & 74.8 & 78.7 & 72.5 & 71.6 & 69.0 & 77.1 & 74.9 & 74.0 \\
w/o Rejection Control   & 73.7 & 75.0 & 78.1 & 72.5 & 75.4 & 74.8 & 77.3 & 74.7 & 75.1 \\
w/o Gating              & 75.1 & 75.8 & 78.3 & 76.4 & 75.8 & 75.1 & 75.8 & 74.1 & 75.8 \\

\midrule
\multicolumn{10}{c}{\textit{(b) Information Density (ID) $\uparrow$}} \\
\midrule
\textbf{ArbGraph (Ours)} & \textbf{247.3} & \textbf{235.5} & \textbf{140.6} & \textbf{241.2} & \textbf{156.4} & \textbf{132.5} & \textbf{124.5} & \textbf{190.2} & \textbf{187.6} \\
w/o Evidence Graph       & 82.1  & 85.4  & 78.6  & 75.2  & 80.3  & 77.8  & 72.4  & 92.6  & 81.4 \\
w/o Soft Arbitration    & 120.1 & 123.8 & 109.9 & 111.3 & 115.3 & 82.3  & 101.4 & 107.9 & 110.0 \\
w/o Initial Screening   & 153.4 & 82.5  & 88.6  & 79.4  & 77.7  & 99.7  & 85.1  & 70.6  & 93.3 \\
w/o Rejection Control   & 97.0  & 185.5 & 112.5 & 238.5 & 79.7  & 130.2 & 73.9  & 160.7 & 132.5 \\
w/o Gating              & 109.2 & 149.3 & 104.4 & 180.0 & 106.8 & 105.5 & 103.5 & 106.9 & 120.7 \\
\bottomrule
\end{tabular}

\label{tab:ablation-results}
\end{table*}

\subsection{Ablation Study}
\label{sec:ablation}

We conduct ablation studies on the LongFact dataset to assess the contribution of key components in ArbGraph.
Table~\ref{tab:ablation-results} reports Fact Recall (FR) and Information Density (ID) across eight domains.
Each variant removes or modifies a single component while keeping other settings fixed.
Overall, removing any major component leads to consistent performance degradation, highlighting the importance of the interaction between structured evidence modeling and iterative arbitration.

\textbf{Evidence graph.}
The \textit{w/o Evidence Graph} variant reduces ArbGraph to a flat set of independent claims.
The substantial performance drop indicates that modeling inter-claim dependencies is critical for capturing both supportive and contradictory relations.
Without an explicit structure, evidence aggregation becomes local and fragmented, preventing the system from forming globally consistent judgments, especially under multi-source or conflicting evidence.

\textbf{Soft conflict arbitration.}
Replacing multi-round arbitration with a single-step strategy (\textit{w/o Soft arbitration}) consistently degrades FR and ID.
This suggests that conflict resolution requires progressive refinement rather than one-shot decisions.
Hard elimination tends to discard partially correct evidence, whereas iterative updates enable gradual calibration of credibility under conflicting signals, leading to more stable evidence integration.

\textbf{Initial relevance screening.}
Removing initial screening (\textit{w/o Initial Screening}) leads to unstable performance.
This reflects the role of early-stage filtering in controlling the search space: without it, irrelevant claims introduce noisy interactions that propagate through the graph and distort subsequent arbitration, especially under dense or weakly related retrieval results.

\textbf{Rejection control.}
Omitting rejection control (\textit{w/o Rejection Control}) reduces Fact Recall, indicating that suppressing low-credibility or contradicted claims is necessary to maintain effective evidence composition.
Without this mechanism, unreliable nodes remain active and interfere with the aggregation of consistent evidence.

\textbf{Binary gating.}
Disabling gating (\textit{w/o Gating}) harms both FR and ID, suggesting that arbitration decisions must be conditioned on sufficient supporting context.
Without confidence-aware updates, unreliable decisions are more likely to propagate, reducing overall stability.

\begin{figure}[t]
    \centering
    \setlength{\tabcolsep}{2pt}
    \renewcommand{\arraystretch}{0.9}
    \begin{tabular}{ccc}
        \includegraphics[width=0.32\columnwidth]{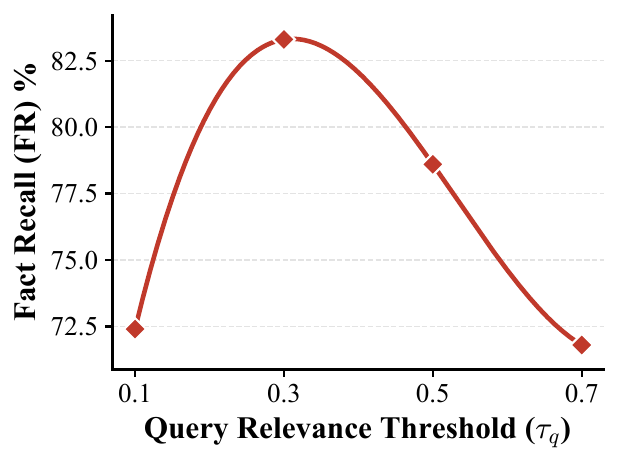} &
        \includegraphics[width=0.32\columnwidth]{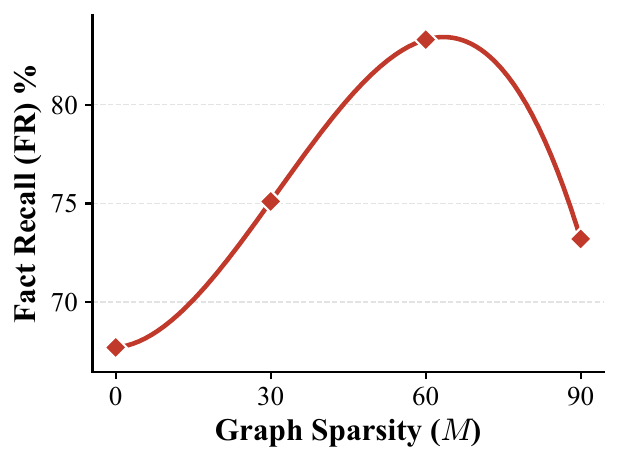} &
        \includegraphics[width=0.32\columnwidth]{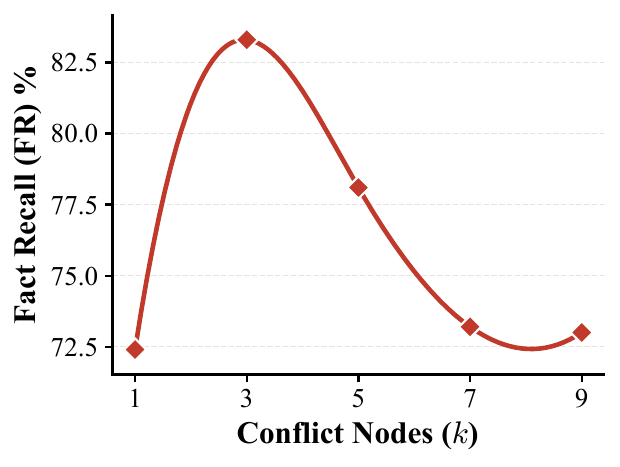} \\
        {\scriptsize (a) FR vs.\ $\tau_q$} &
        {\scriptsize (c) FR vs.\ $M$} &
        {\scriptsize (e) FR vs.\ $k$} \\

        \includegraphics[width=0.32\columnwidth]{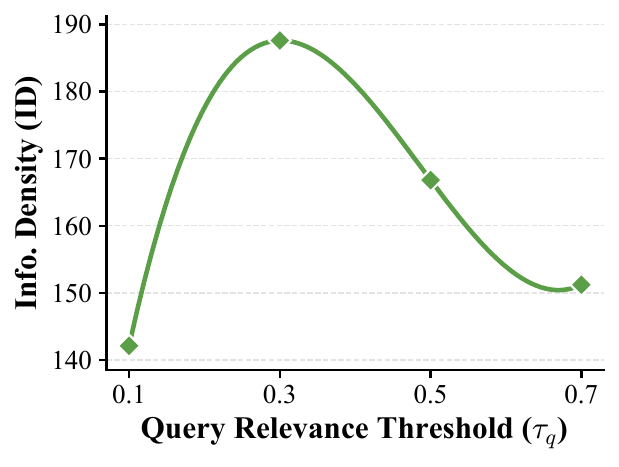} &
        \includegraphics[width=0.32\columnwidth]{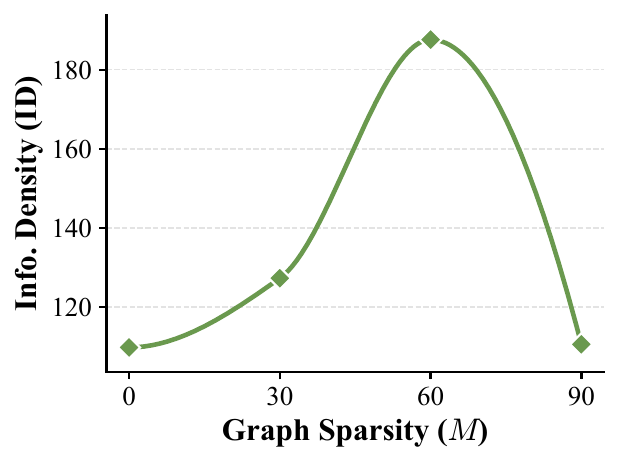} &
        \includegraphics[width=0.32\columnwidth]{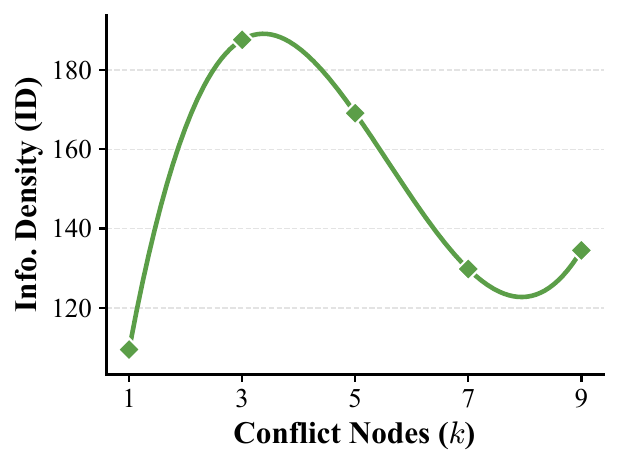} \\
        {\scriptsize (b) ID vs.\ $\tau_q$} &
        {\scriptsize (d) ID vs.\ $M$} &
        {\scriptsize (f) ID vs.\ $k$}
    \end{tabular}
    \caption{\small \textbf{Sensitivity analysis of key parameters.}
    Moderate values of $\tau_q$, $M$, and $k$ achieve a balance between evidence coverage, structural connectivity, and arbitration stability.}
    \label{fig:all_sensitivity}
\end{figure}

\subsection{Analysis of Graph Construction Factors}

We conduct the following sensitivity analyses on the LongFact benchmark,
where claim-level annotations enable fine-grained evaluation of how graph construction
affects factual coverage and evidence aggregation.

\subsubsection{Sensitivity to Query Relevance Threshold ($\tau_q$)}

We further examine the sensitivity of ArbGraph to the query relevance threshold $\tau_q$, which controls whether a claim node is retained based on its semantic similarity to the query via the query-aware filtering criterion in Eq.~\ref{eq:query-aware-filter}.
Figure~\ref{fig:all_sensitivity}(a) and (b) report the effects of varying $\tau_q$ on Fact Recall (FR) and Information Density (ID).

When $\tau_q$ is set too low, a large number of weakly related background claims are introduced into the evidence graph.
Although coverage increases, excessive noise dilutes informative content and interferes with evidence arbitration, leading to degraded ID and even reduced FR.
This behavior reflects the dependence of arbitration on the quality of initial node selection, as noisy nodes may dominate local interactions and distort credibility propagation.

In contrast, overly strict thresholds aggressively prune claim nodes, discarding supporting but implicitly relevant evidence and causing a substantial drop in recall.
Such pruning reduces the diversity of evidence sources, limiting the ability of the graph to capture complementary support signals.

A moderate threshold ($\tau_q=0.3$) achieves the best balance between relevance and coverage, yielding the highest FR and ID.
Accordingly, we adopt $\tau_q=0.3$ in all subsequent experiments.
This result highlights that effective graph construction requires a careful balance between noise suppression and evidence completeness.

\subsubsection{Sensitivity to Graph Sparsity ($M$)}
\label{sec:M_sensitivity}

We analyze the sensitivity of \textit{ArbGraph} to the graph sparsity parameter $M$, which controls the maximum number of support edges retained after relation verification.
Specifically, all contradiction edges are preserved, while only the top-$M$ support edges ranked by semantic similarity are kept.

As shown in Figure~\ref{fig:all_sensitivity}(c) and (d), the results exhibit a clear bell-shaped trend.
When $M$ is small ($M=0$ or $30$), performance is limited by insufficient connectivity, leading to a fragmented graph where support and contradiction signals cannot effectively propagate.

Performance peaks at $M=60$, achieving the highest Fact Recall (83.3\%) and Information Density (187.6), indicating that moderate sparsity balances noise filtering and relational coverage.
In contrast, larger values ($M=90$) introduce noisy or weakly related edges, which dilute meaningful constraints and reduce stability.

Overall, retaining a moderate number of support edges (Top-$60$) provides a robust trade-off between structural efficiency and reasoning effectiveness.

\subsubsection{Discussion on Semantic Similarity Threshold ($\tau_{\text{sim}}$)}

The similarity threshold $\tau_{\text{sim}}$ controls the semantic neighborhood for candidate pair mining.
A lower threshold increases graph density but may introduce spurious relations and amplify noise,
while a higher threshold enforces sparsity at the risk of missing valid interactions.

Empirically, ArbGraph remains stable within a moderate range of $\tau_{\text{sim}}$, with only limited performance variation observed.
This robustness arises because arbitration operates over aggregated multi-source evidence, rather than relying on individual pairwise relations.
As a result, occasional spurious or missing edges have limited impact on the overall arbitration outcome.

These observations suggest that effective graph construction requires balancing relational coverage with structural reliability, ensuring that informative interactions are preserved while avoiding excessive noise.

\subsubsection{Sensitivity to Conflict Intensity ($k$)}
\label{sec:k_sensitivity}

We analyze the sensitivity of \textit{ArbGraph} to the conflict intensity $k$, which controls the number of contradiction pairs prioritized for arbitration in each round.

Figure~\ref{fig:all_sensitivity}(e) and (f) show that increasing $k$ from small to moderate values improves both Fact Recall (FR) and Information Density (ID), as a larger arbitration budget enables better reconciliation of locally inconsistent evidence.

However, performance degrades when $k$ becomes too large, suggesting that overly aggressive arbitration introduces noise by forcing decisions on marginal or weakly supported contradictions.

Overall, a moderate value of $k$ provides the best trade-off, supporting stable performance and indicating that resolving high-impact conflicts is more effective than exhaustive arbitration.

\subsubsection{Cost-Performance Trade-off Analysis}
\label{sec:trade-off}

A recurring pattern across our sensitivity analyses for $\tau_q$, $M$, and $k$ is the emergence of a clear bell-shaped trend (see Figure~\ref{fig:all_sensitivity}). This pattern provides insight into the relationship between computational investment and factual robustness.

Specifically, increasing the graph sparsity parameter $M$ or the arbitration budget $k$ introduces additional computation (e.g., more relation verification or arbitration steps). However, beyond moderate values, performance degrades, suggesting that excessive computation introduces structural or reasoning noise (e.g., spurious edges or marginal contradictions) that dilutes informative evidence.

In contrast, ArbGraph performs best under moderate settings (e.g., $\tau_q=0.3$, $M=60$, $k=3$), effectively resolving high-impact conflicts while maintaining high information density. Overall, these results indicate that ArbGraph does not exhibit monotonic scaling with respect to computational cost, and that optimal performance depends on balanced, structure-aware allocation of computation.
\begin{figure}[t]
  \centering
  \begin{minipage}[t]{0.40\columnwidth}
    \centering
    \includegraphics[width=\linewidth]{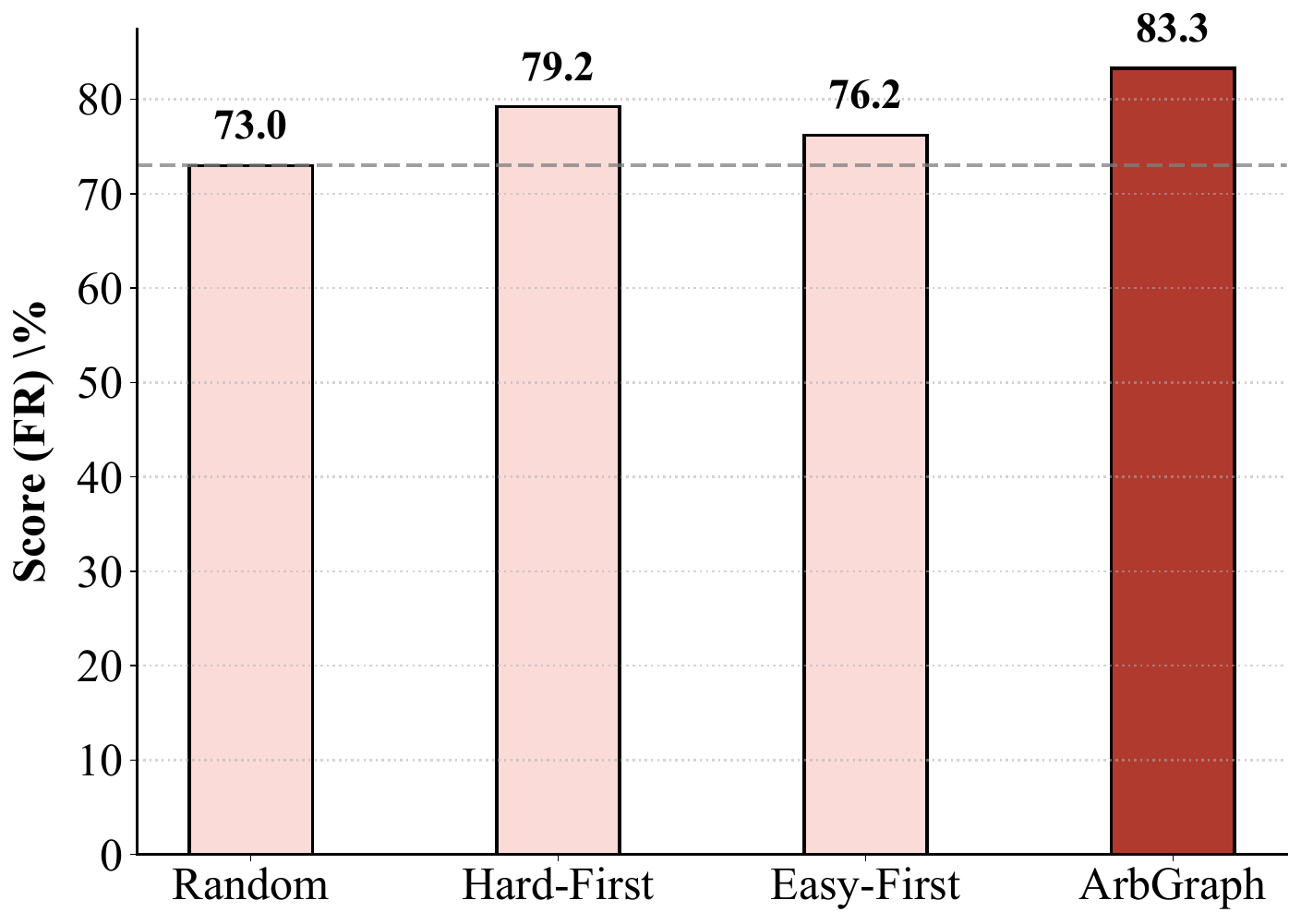}
    
    {\scriptsize (a) Fact Recall (FR)}
  \end{minipage}
  \hspace{0.06\columnwidth}
  \begin{minipage}[t]{0.40\columnwidth}
    \centering
    \includegraphics[width=\linewidth]{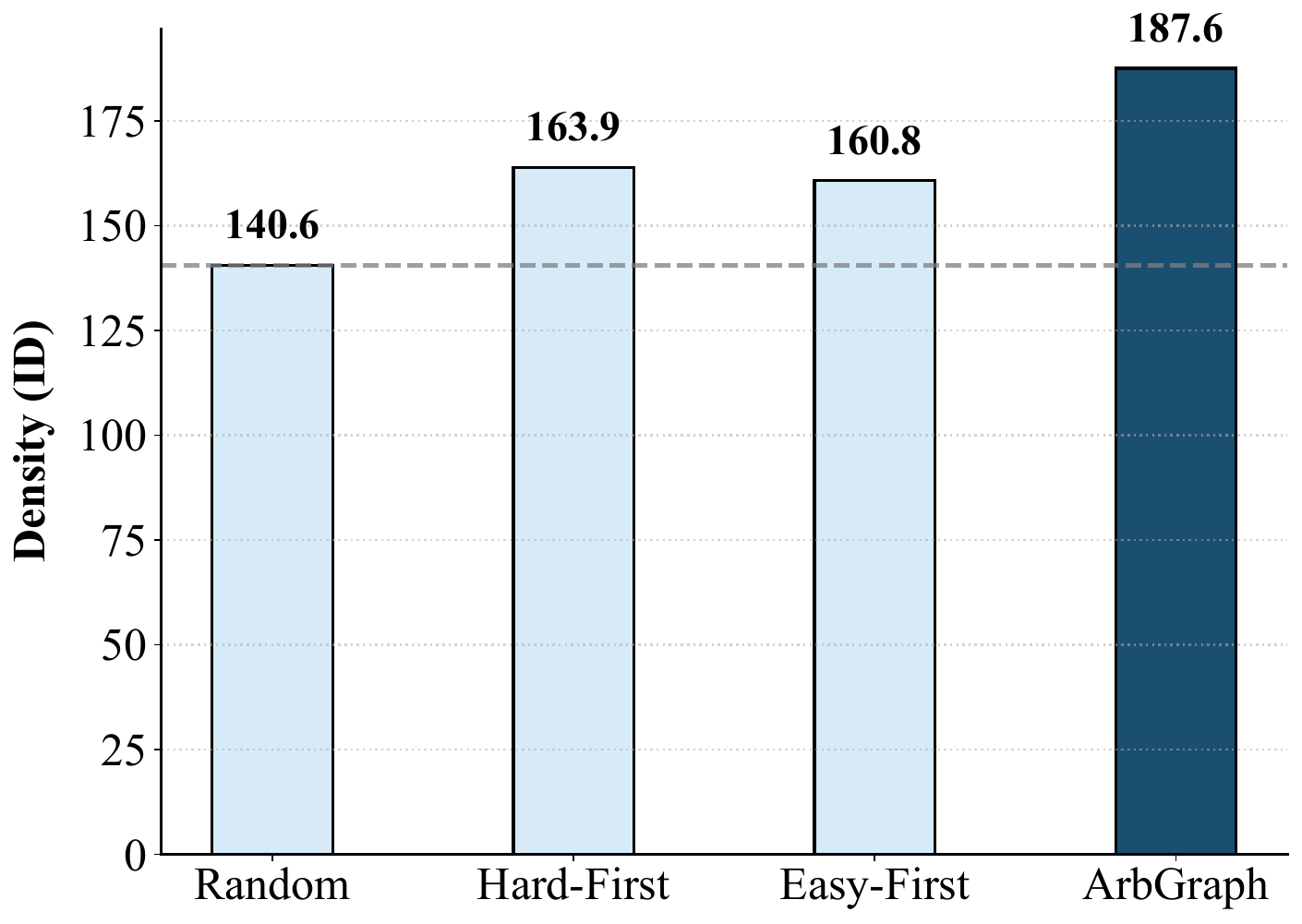}
    
    {\scriptsize (b) Information Density (ID)}
  \end{minipage}
  \caption{\small \textbf{Comparison of Scheduling Policies.}
  ArbGraph’s conflict-aware scheduling improves factuality and evidence density.}
  \label{fig:main_performance}
\end{figure}

\subsubsection{Effectiveness of Conflict-Aware Scheduling}
\label{sec:scheduling_analysis}

We analyze the impact of different scheduling strategies on the arbitration process.
While heuristic-based scheduling policies such as \textit{Hard-First} and \textit{Easy-First} rely on static ordering rules,
ArbGraph employs a dynamic, conflict-aware scheduling policy guided by the \textit{Intensity Score} (Eq.~\ref{eq:intensity-score}) within the evidence graph.
This dynamic mechanism further enables adaptive allocation of reasoning effort toward high-impact conflicts, while continuously adjusting the arbitration trajectory based on evolving graph states, rather than following a fixed processing order.

As shown in Figure~\ref{fig:main_performance}, ArbGraph consistently outperforms all heuristic-based alternatives.
Compared with static scheduling strategies, conflict-aware scheduling prioritizes claim pairs with higher contradiction intensity,
allowing the model to resolve the most critical inconsistencies earlier in the arbitration process.
This leads to more stable intermediate states and reduces the propagation of unresolved conflicts.

In particular, ArbGraph achieves higher Fact Recall (FR) and the best Information Density (ID) among all compared strategies.
This indicates that effective scheduling is not merely about processing order, but about selectively focusing on structurally important conflicts.
Overall, these results highlight that leveraging graph-based conflict signals is essential for efficient and reliable evidence arbitration.

\subsection{Qualitative and Reliability Analysis}

\subsubsection{Component-wise Reliability Analysis and Error Propagation}
\label{sec:reliability}

We conduct a component-wise reliability analysis of ArbGraph’s key modules, including relation verification and arbitration.

For relation verification, we construct a manually annotated set of 200 claim pairs from LongFact, labeled as Support, Contradict, or Neutral. As shown in Table~\ref{tab:reliability}, the verifier achieves high accuracy, indicating a generally reliable evidence graph. For arbitration, evaluation on 50 representative cases yields 96.0\% accuracy, indicating stable decisions under noisy retrieval.

We make three observations on the impact of local errors. First, structural redundancy across multiple evidence sources mitigates isolated verification errors (approximately $\sim$4\%). Second, mechanism-level safeguards, including query-aware filtering and confidence-based gating, prevent low-confidence updates from propagating. Third, manual tracing shows that incorrect intermediate updates are often filtered by the acceptance threshold ($\tau_{accept}$) or not selected during generation.

Overall, these results suggest that ArbGraph is robust to local errors, as inaccuracies are effectively contained by the graph structure and arbitration mechanisms, though more comprehensive evaluation remains future work.

\begin{table}[t]
\centering
\small
\setlength{\tabcolsep}{6pt}
\caption{Component-wise reliability analysis of ArbGraph.}
\begin{tabular}{lccc}
\toprule
Component & Category & Samples & Accuracy \\
\midrule
Relation Verifier & Support    & 72  & 95.8\% \\
                  & Contradict & 51  & 96.1\% \\
                  & Neutral    & 77  & 96.1\% \\
                  & \textit{Avg.} & 200 & 96.5\% \\
\midrule
Conflict Arbitrator & Arbitration & 50 & 96.0\% \\
\bottomrule
\end{tabular}
\label{tab:reliability}
\end{table}

\begin{figure*}[t]
\centering
\footnotesize
\setlength{\tabcolsep}{5pt}
\renewcommand{\arraystretch}{1.2}

\setlength{\aboverulesep}{0pt}
\setlength{\belowrulesep}{0pt}

\begin{tabular}{p{0.47\linewidth} | p{0.47\linewidth}}
\toprule
\multicolumn{2}{c}{\textbf{(A) Input: Query \& Retrieved Documents}} \\
\midrule
\multicolumn{2}{p{0.96\linewidth}}{
\vspace{2pt}
\textbf{Query:} \textit{``How is the United States related to the East Asia Summit (EAS)?''} \newline
\textbf{Retrieved documents:} 
(D1) \textit{EAS Wiki}: U.S. joined in 2011. \quad
(D2) \textit{ASEAN 2025}: U.S. attended meetings. \quad
(D3) \textit{Trump Trips}: Skipped 2017 meeting. \quad
(D4) \textit{Homonym}: Emergency Alert System (EAS).
\vspace{2pt}
} \\
\midrule
\multicolumn{1}{c|}{\textbf{(B) Standard RAG: Flat \& Vulnerable}} & 
\multicolumn{1}{c}{\textbf{(C) Ours: Evidence-Level Arbitration}} \\
\midrule

\begin{minipage}[t]{\linewidth}
\vspace{4pt}
\textbf{Step 0: Input processing.}  
The model ingests all retrieved texts equally.

\textbf{Step 1: Attention dilution.}
\begin{itemize}
  \setlength{\itemsep}{0pt}
  \setlength{\parskip}{0pt}
  \setlength{\parsep}{0pt}
  \item \textbf{Homonym noise:} ``Emergency Alert System'' (D4).
  \item \textbf{Fact decay:} ``joined 2011'' weakened.
\end{itemize}

\textbf{Step 2: Implicit conflict.}
\begin{itemize}
  \setlength{\itemsep}{0pt}
  \item ``Skipping'' misread as non-membership.
\end{itemize}

\textbf{Outcome (\textcolor{red}{failure}):} \\
\textit{``The U.S. is not a member of the EAS.''}
\vspace{4pt}
\end{minipage}
&
\begin{minipage}[t]{\linewidth}
\vspace{4pt}
\textbf{Step 0: Claim extraction.}
\begin{itemize}
  \setlength{\itemsep}{0pt}
  \setlength{\parskip}{0pt}
  \setlength{\parsep}{0pt}
  \item $c_1$: joined EAS \quad $c_3$: skipped meeting
  \item $c_2$: attended meetings
  \item \textcolor{gray}{$c_4$: alert system}
\end{itemize}

\textbf{Step 1: Filtering.}
\begin{itemize}
  \setlength{\itemsep}{0pt}
  \item \textcolor{gray}{$c_4$ discarded}, \textcolor{gray}{$c_2$ deprioritized}
  \item \textbf{Retained:} $c_1$, $c_3$
\end{itemize}

\textbf{Step 2: Arbitration.}
\begin{itemize}
  \setlength{\itemsep}{0pt}
  \item membership and attendance are distinct relations
\end{itemize}

\textbf{Outcome (\textcolor{blue}{success}):} \\
\textit{``The U.S. joined the EAS in 2011.''}
\vspace{4pt}
\end{minipage}
\\
\bottomrule
\end{tabular}

\caption{\textbf{Comparison of standard RAG vs. ArbGraph.} Standard RAG fails due to noise and unresolved conflicts, while ArbGraph resolves conflicts via evidence-level arbitration.}
\label{fig:eas_case_study}
\end{figure*}

\subsubsection{Case Study}
\label{sec:case-study}

We analyze a challenging query: \textit{``How is the United States related to the East Asia Summit (EAS)?''} (Figure~\ref{fig:eas_case_study}), which requires distinguishing formal membership from episodic attendance under ambiguous evidence.

The retrieved documents contain the target fact ($D_1$: U.S. joined in 2011) mixed with heterogeneous noise, including contextual ambiguity from attendance-related descriptions ($D_2, D_3$) and homonym noise ($D_4$) referring to the unrelated ``Emergency Alert System.'' Notably, these signals correspond to different semantic roles, making naive aggregation prone to confusion.

Standard RAG (Panel~B) processes documents as a flat sequence and fails to resolve these ambiguities. The presence of homonym noise dilutes attention, and the implicit conflict between ``joining'' ($D_1$) and ``skipping'' ($D_3$) is misinterpreted, leading to the incorrect conclusion that the U.S. is not a member of the EAS. Without explicit modeling of such conflicts, contradictory evidence is implicitly treated as complementary, amplifying reasoning errors.

In contrast, ArbGraph (Panel~C) decomposes documents into atomic claims and filters structurally irrelevant nodes. It then identifies and resolves the conflict between status and attendance claims through evidence-level arbitration. This explicit separation between status and event-level evidence avoids spurious interactions during reasoning. Conditioning generation on this consistent structure enables ArbGraph to correctly infer formal membership.

\subsection{Implications for Long-Form RAG Design}
\label{sec:design-insights}

Based on the empirical results on LongFact and RAGChecker, we derive several key insights into the design of reliable long-form RAG systems.

\textbf{Evidence consistency is more critical than retrieval breadth.}
While aggressive retrieval expansion may improve recall, it also introduces irrelevant or contradictory evidence.
Our results show that increased coverage does not necessarily translate into better generation quality, as noisy evidence can degrade both factual density and reliability.
In contrast, ArbGraph achieves strong performance by operating on a smaller but internally consistent set of verified claims, highlighting the importance of evidence coherence over sheer quantity.

\textbf{Pre-generation arbitration is more robust than post-hoc correction.}
Post-hoc methods are limited by the \textit{error cascade}: once incorrect information is introduced, it propagates through the generation process and becomes exceedingly difficult to correct.
By resolving conflicts before decoding, ArbGraph ensures that the model is conditioned on a coherent evidence set, thereby reducing error propagation and improving output stability.

\textbf{Structured evidence modeling enables explicit conflict resolution.}
Unlike prior graph-based RAG approaches that primarily focus on retrieval connectivity, ArbGraph uses graphs to explicitly model \textit{support} and \textit{contradiction} relations.
This transforms evidence integration from implicit, token-level aggregation into structured reasoning over competing claims, enabling explicit credibility estimation prior to generation.

\textbf{Improved reliability comes with additional pre-generation cost.}
ArbGraph introduces extra steps such as claim extraction, relation verification, and iterative arbitration, which invariably increase computational cost and latency.
However, this overhead reflects a necessary shift from generation-time heuristics to structured pre-generation reasoning.
By resolving high-impact conflicts early, ArbGraph secures a highly dependable foundation for downstream generation, highlighting a fundamental trade-off between factual reliability and computational efficiency.

Overall, these findings suggest that reliable long-form RAG requires not only stronger retrieval mechanisms and generation models, but also the effective organization and arbitration of evidence prior to synthesis.
\section{Conclusion and Future Work}
\label{sec:conclusion}

In this paper, we presented \textbf{ArbGraph}, an evidence-level arbitration framework designed to address the persistent challenge of factual inconsistencies in long-form retrieval-augmented generation. 
By shifting from the conventional implicit integration of retrieved content to a structured {``arbitrate-first, generate-later''} paradigm, ArbGraph establishes a more reliable factual foundation for the generation process. 
The framework's core strength lies in its ability to decompose evidence into atomic claims and model their logical interdependencies via a contradiction-aware evidence graph. 
Experimental results on LongFact and RAGChecker demonstrate that ArbGraph consistently outperforms strong baselines in terms of factual coverage, grounding quality, and robustness under noisy or conflicting retrieval conditions. 
These findings highlight the importance of explicitly modeling and resolving evidence-level conflicts as a principled approach toward trustworthy long-form RAG.

Looking ahead, several directions are worth further exploration to improve the practicality and generality of the proposed framework. 
First, the current evidence graph focuses on support and contradiction relations; extending it to capture richer logical structures, such as temporal dependencies or causal relations, may further enhance reasoning capabilities. 
Second, while the iterative arbitration process improves evidence reliability, it introduces additional pre-generation reasoning overhead. 
Understanding the trade-off between factual consistency and computational efficiency, as well as developing lightweight arbitrators or adaptive scheduling strategies, remains an important direction for future work. 
Finally, we plan to extend ArbGraph to multimodal settings, enabling consistent reasoning across heterogeneous evidence sources (e.g., text and images), and to explore its integration with post-hoc verification methods to form a unified multi-stage reliability framework.






\end{document}